\documentclass[11pt]{article}

\usepackage[final]{acl}

\usepackage{times}
\usepackage{latexsym}
\usepackage{pifont}

\usepackage[T1]{fontenc}
\usepackage[utf8]{inputenc}

\usepackage{microtype}

\usepackage{inconsolata}

\usepackage{graphicx}

\usepackage{booktabs}

\usepackage{colortbl}
\usepackage{xcolor}  

\usepackage{amsmath,amssymb,amsfonts,mathtools}
\allowdisplaybreaks
\DeclareMathOperator{\KL}{KL}
\usepackage[most]{tcolorbox}

\title{Rethinking the Text-Vision Reasoning Imbalance in MLLMs through the Lens of Training Recipes}

\author{
Guanyu Yao$^{1}$\thanks{~~Equal contribution.},
Qiucheng Wu$^{1}$\footnotemark[1],
Yang Zhang$^{2}$,
Zhaowen Wang$^{3}$,
Handong Zhao$^{3}$,
Shiyu Chang$^{1}$ \\
$^{1}$UC Santa Barbara \quad
$^{2}$MIT-IBM Watson AI Lab \quad
$^{3}$Adobe Research \\
\texttt{\{gyao,qiucheng,chang87\}@ucsb.edu} \quad
\texttt{yang.zhang2@ibm.com} \quad
\texttt{\{zhawang,hazhao\}@adobe.com} \\
}

\begin{document}
\maketitle
\begin{abstract}
Multimodal large language models (MLLMs) have demonstrated strong capabilities on vision-and-language tasks. However, recent findings reveal an imbalance in their reasoning capabilities across visual and textual modalities. Specifically, current MLLMs often over-rely on textual cues while under-attending to visual content, resulting in suboptimal performance on tasks that require genuine visual reasoning. We refer to this phenomenon as the \textit{modality gap}, defined as the performance disparity between text-centric and vision-centric inputs. In this paper, we analyze the modality gap through the lens of training recipes. We first show that existing training recipes tend to amplify this gap. Then, we systematically explore strategies to bridge it from two complementary perspectives: data and loss design. Our findings provide insights into developing training recipes that mitigate the modality gap and promote a more balanced multimodal reasoning.
\end{abstract}

% Multimodal large language models (MLLMs) have shown exceptional capability in vision-and-language tasks. 

\section{Introduction}
\label{sec:intro}
\begin{figure}[t]
    \centering
    \includegraphics[width=1.0\columnwidth]{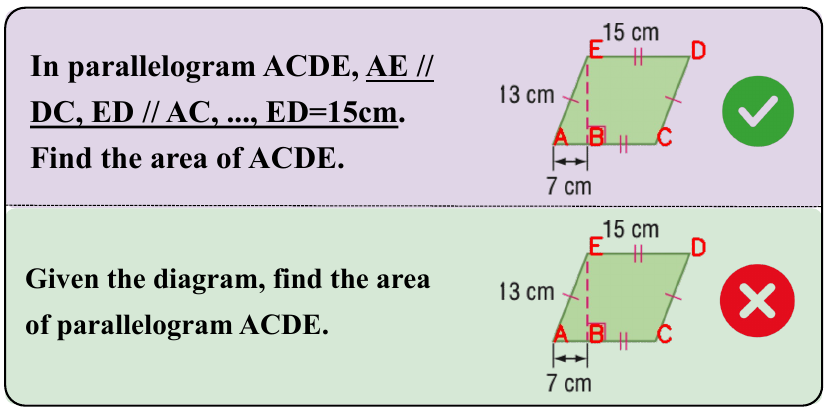}
    \caption{Current MLLMs exhibit an imbalance between visual and textual reasoning. When information present in the visual modality is removed from the text, the MLLM fails to answer the question.}
    \label{fig:intro}
    \vspace{-0.15in}
\end{figure}

Multimodal large language models (MLLMs) have shown exceptional reasoning capabilities on complex tasks that require multimodal reasoning. However, recent studies~\cite{zhang2024mathverse,li2025vista} reveal a reasoning imbalance: these models often rely heavily on textual cues while under-exploiting visual information when generating answers. This over-reliance on text leads to suboptimal results on tasks that require genuine visual reasoning. We refer to this phenomenon as the \textit{modality gap}. As exemplified in Figure~\ref{fig:intro}, when critical information present in the visual modality is removed from the text, MLLMs fail to answer questions that could have been correctly answered when the full text was provided, highlighting their insufficient visual reasoning.

To understand the origin of this imbalance, we focus on the training recipes of current MLLMs. An important observation is that many training samples contain overlapping information across the textual and visual modalities. In such cases, it may be easier for MLLMs to rely on the already complete textual information rather than engage in visual reasoning. We hypothesize that this training process largely contributes to the modality gap. Our preliminary evidence supports this view: under standard training setups, the gap between text- and vision-centric performance widens over time, underscoring the need for more balanced training strategies.

Building on these insights, our goal is to identify improved training recipes for MLLMs that jointly \ding{172} ensure the effective use of visual information and \ding{173} maintain or enhance overall reasoning ability in the targeted domains. We approach this problem from two perspectives: \textit{data} and \textit{loss}.
From the data perspective, we consider vision-centric and text-centric data in supporting balanced multimodal reasoning and explore how simple data mixing and carefully designed curriculum strategies can leverage the strengths of both modalities.
From the loss perspective, we propose a KL-based self-distillation objective that transfers reasoning competence from full-text to partial-text inputs, preserving general reasoning performance while strengthening visual grounding.
Our key contributions are as follows:
\begin{itemize}
    \item We establish a diagnostic that reveals a consistent discrepancy between text-centric and vision-centric performance across a range of public and private MLLMs at different scales. Furthermore, we show that current training setups \textit{widen} this discrepancy, underscoring the need for targeted recipes beyond naïve RL.
    \item We propose improved RL training recipes from the perspectives of \emph{data} and \emph{loss}, designed to preserve reasoning competence in the textual modality while reducing the reasoning gap across modalities.
\end{itemize}

\section{Related Work}
\label{sec:related}
\subsection{Multimodal Large Language Models}
Multimodal Large Language Models (MLLMs) have emerged as powerful tools that integrate visual and textual information to perform a wide range of tasks, including image captioning, visual question answering, and geometric reasoning. Notable MLLMs include Qwen2.5-VL series~\cite{Qwen2.5-VL,Qwen2-VL,Qwen-VL}, VL-Rethinker series~\cite{wang2025vl}, MiniCPM series~\cite{yao2024minicpm}, InternVL3.5 series~\cite{wang2025internvl3}, Kimi-VL series~\cite{team2025kimi}, Gemma series~\cite{team2025gemma}, GPT-5~\cite{openai2025gpt5} and Gemini~\cite{comanici2025gemini}. These models typically employ a combination of pre-trained vision encoders and large language models, fine-tuned on multimodal datasets to enhance their understanding and generation capabilities across both modalities.

\subsection{Visual Reasoning in MLLMs}
Visual reasoning is a critical capability for MLLMs, enabling them to interpret and reason about visual content in conjunction with textual information. Recent studies like MathVerse~\cite{zhang2024mathverse} have highlighted the challenges MLLMs face in effectively utilizing visual information, often defaulting to text-based cues. This has led to the identification of the modality gap, where models perform significantly better on text-centric tasks compared to vision-centric ones. 

To address this issue, various approaches have been proposed, including specialized training datasets~\cite{liu2024llavanext,li2024llava,gao2023g}, model architecture design~\cite{lu2024ovis,bigverdi2025perception}, and loss functions that encourage visual attention~\cite{luo2024deem,li2025vista,wang2025perception}. In this paper, we build upon these foundations by exploring RL-based methods to enhance visual reasoning while mitigating the modality gap.

\begin{table}[t]
    \centering
    \resizebox{\columnwidth}{!}{
    \begin{tabular}{l l ccc c}
    \toprule
    \textbf{Model} & \textbf{Dataset} & \textbf{Text} & \textbf{Vision} & \textbf{Avg} & \cellcolor{gray!20}\textbf{Gap} \\
    \midrule
    Qwen2.5-VL 3B & PGPS9K & \textit{23.97} & \textit{18.12} & \textit{21.05} & \cellcolor{gray!20}\textbf{\textit{5.85}} \\
     & MathVerse & \textit{35.68} & \textit{28.66} & \textit{31.47} & \cellcolor{gray!20}\textit{7.02} \\
    Qwen2.5-VL 7B & PGPS9K & \textit{37.75} & \textit{29.98} & \textit{33.87} & \cellcolor{gray!20}\textit{7.77} \\
     & MathVerse & \textit{55.01} & \textit{45.76} & \textit{51.10} & \cellcolor{gray!20}\textit{9.25} \\
    MiniCPM-V-4 & PGPS9K & \textit{34.70} & \textit{30.20} & \textit{32.45} & \cellcolor{gray!20}\textit{4.50} \\
     & MathVerse & \textit{44.35} & \textit{37.21} & \textit{40.07} & \cellcolor{gray!20}\textit{7.14} \\
    Gemma-3-4b-it & PGPS9K & \textit{40.50} & \textit{26.12} & \textit{35.59} & \cellcolor{gray!20}\textit{14.38} \\
     & MathVerse & \textit{42.36} & \textit{33.50} & \textit{37.05} & \cellcolor{gray!20}\textit{8.86} \\
    Kimi-VL-A3B & PGPS9K & \textit{40.57} & \textit{31.32} & \textit{35.95} & \cellcolor{gray!20}\textit{9.25} \\
     & MathVerse & \textit{57.91} & \textit{48.27} & \textit{51.23} & \cellcolor{gray!20}\textit{9.64} \\
    VL-Rethinker 7B & PGPS9K & \textit{40.45} & \textit{36.05} & \textit{38.25} & \cellcolor{gray!20}\textit{4.40} \\
     & MathVerse & \textit{65.42} & \textit{57.28} & \textit{60.53} & \cellcolor{gray!20}\textit{8.14} \\
    InternVL3.5 8B & PGPS9K & \textit{52.18} & \textit{39.65} & \textit{45.92} & \cellcolor{gray!20}\textit{12.53} \\
     & MathVerse & \textit{66.68} & \textit{54.19} & \textit{59.19} & \cellcolor{gray!20}\textit{12.49} \\
    Qwen3-VL & PGPS9K & \textit{69.70} & \textit{66.99} & \textit{68.35} & \cellcolor{gray!20}\textit{2.71} \\
     & MathVerse & \textit{64.06} & \textit{60.89} & \textit{61.67} & \cellcolor{gray!20}\textbf{\textit{3.17}} \\
    \midrule
    GPT-5\footnotemark[1] & PGPS9K & \textbf{\textit{94.00}} & \textbf{\textit{80.00}} & \textbf{\textit{87.00}} & \cellcolor{gray!20}\textit{14.00} \\
     & MathVerse & \textbf{\textit{76.67}} & \textit{63.33} & \textit{70.00} & \cellcolor{gray!20}\textit{13.34} \\
    Gemini 2.5 Flash\footnotemark[1] & PGPS9K & \textit{92.00} & \textit{74.00} & \textit{83.00} & \cellcolor{gray!20}\textit{18.00} \\
     & MathVerse & \textit{86.96} & \textbf{\textit{77.78}} & \textbf{\textit{82.00}} & \cellcolor{gray!20}\textit{9.18} \\
    \bottomrule
    \end{tabular}}
    \caption{Base model performance.}
    \label{tab:baseline_results_combined}
    \vspace{-0.1in}
\end{table}
\footnotetext[1]{Due to API costs, the results are evaluated on a subset of 50 test samples.}

\section{Modality Gap in MLLMs}

We begin by quantifying the modality gap across a range of open-source and commercial MLLMs. To illustrate this gap, we consider two kinds of data:

\noindent $\bullet$ $\mathcal{D}_1$: \textbf{Text-centric.} All necessary information is contained within the provided text, and the MLLM can solve the problem through textual reasoning.

\noindent $\bullet$ $\mathcal{D}_2$: \textbf{Vision-centric.} Some necessary information is present in the image but not in the text, requiring the MLLM to perform visual reasoning to successfully solve the problem.

To construct $\mathcal{D}_1$ and $\mathcal{D}_2$, we draw upon two challenging visual reasoning datasets: PGPS9K \cite{zhang2023multi} and MathVerse \cite{zhang2024mathverse}. In PGPS9K, each question consists of a textual condition and a question statement, accompanied by a fully annotated figure that specifies entities and their relations. Accordingly, we define $\mathcal{D}_1$ as the setting where both the image and text provide complete information, and $\mathcal{D}_2$ as the setting where information present in the image has been removed from the text.
For MathVerse, following prior work, we define $\mathcal{D}_1$ and $\mathcal{D}_2$ subsets to focus respectively on text (\emph{Text-Dominant} and \emph{Text-Lite} subsets) and vision (\emph{Vision-Intensive}, \emph{Vision-Dominant}, and \emph{Vision-Only} subsets) reasoning capabilities. Further details of the datasets are provided in Appendix~\ref{appendix:dataset-details}.

\paragraph{Metrics.}
We report the \textbf{text-centric} and \textbf{vision-centric} performance measured on $\mathcal{D}_1$ and $\mathcal{D}_2$. In addition, we report the \textbf{overall} performance as the average accuracy across $\mathcal{D}_1$ and $\mathcal{D}_2$.

\paragraph{Direct Inference Results.} We begin by evaluating a series of off-the-shelf MLLMs.
The results are summarized in Table~\ref{tab:baseline_results_combined}. Across both the PGPS9K and MathVerse datasets, we observe a consistent modality gap: text-centric performance is consistently higher than vision-centric performance across various open-source and commercial models of different sizes. Moreover, stronger MLLMs tend to exhibit a larger performance gap. This discrepancy underscores the need for targeted strategies to enhance the visual reasoning capabilities of MLLMs.

\paragraph{Effect of Standard RL Training.}Next, we explore how standard training influences the modality gap. 
In this experiment, we apply DAPO~\cite{yu2025dapo} to fine-tune Qwen2.5-VL (3B and 7B) under both $\mathcal{D}_1$ and $\mathcal{D}_2$ settings from the PGPS9K training set.
Note that all figures in $\mathcal{D}_1$ and $\mathcal{D}_2$ have their entities, relations, and other geometric properties explicitly annotated.
Thus, the model can always obtain complete information related to the question from the image.
As shown in Table~\ref{tab:simple_rl_results_combined}, training on $\mathcal{D}_1$ primarily improves text-centric performance but enlarges the modality gap as training progresses, whereas training on $\mathcal{D}_2$ strengthens vision-centric performance and narrows the gap, though at the expense of overall accuracy.

\begin{figure}[t]
    \vspace{-0.05in}
    \centering
    \includegraphics[width=\columnwidth]{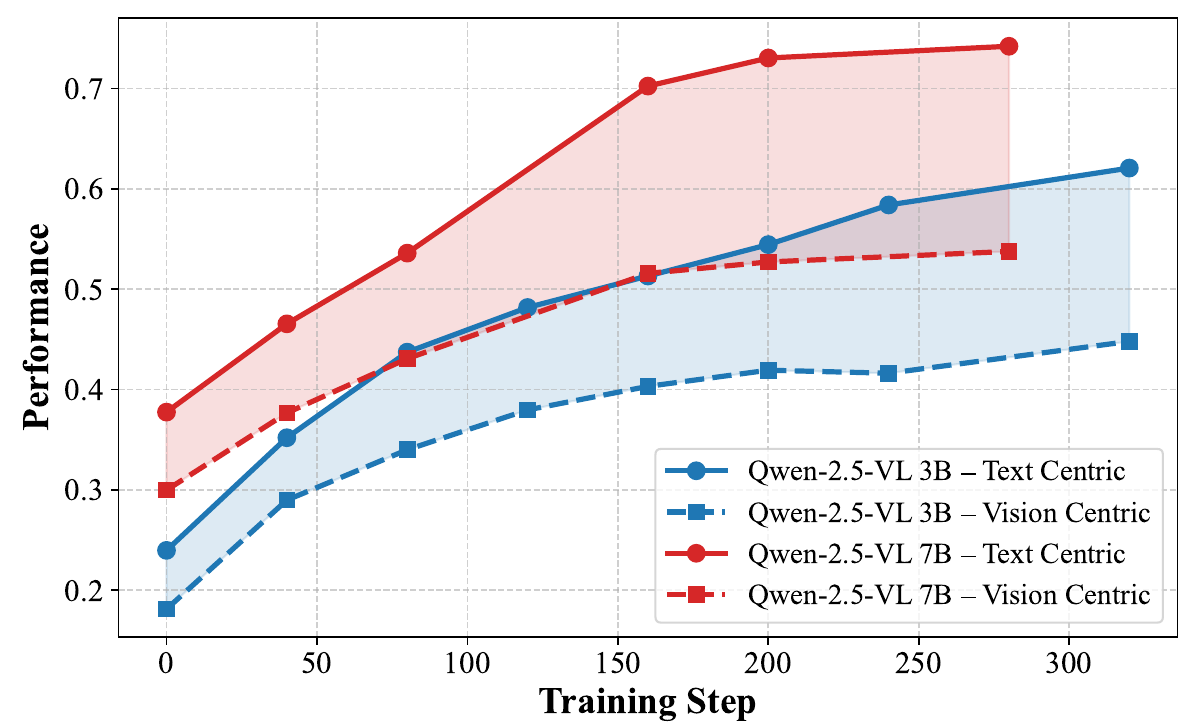}
    \vspace{-0.24in}
    \caption{Standard training recipe widens modality gap.}
    \label{fig:gap-in-training}
    \vspace{-0.08in}
\end{figure}

Moreover, as shown in Figure~\ref{fig:gap-in-training}, during standard training on $\mathcal{D}_1$, the modality gap progressively increases with training steps.
These observations indicate that the standard training recipe is insufficient to resolve the modality gap in MLLMs, highlighting the need for more nuanced training strategies.

\begin{figure*}
\centering
\vspace{-0.5em}
\includegraphics[width=\textwidth]{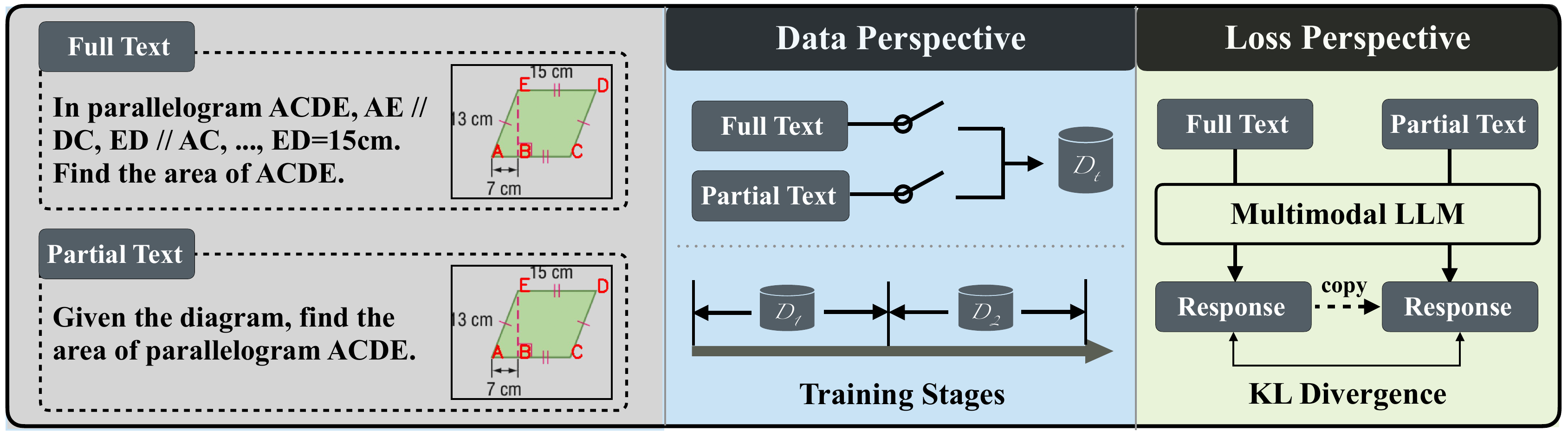}
\caption{We consider two types of data: \ding{182} both text and image contain complete information, referred to as \textit{full text}; and \ding{183} the text omits information already present in the image, referred to as \textit{partial text}. We then analyze better training recipe from both data and loss perspectives.}
\label{fig:method01}
\vspace{-7pt}
\end{figure*}

\begin{table}[t]
    \centering
    \resizebox{\columnwidth}{!}{
    \begin{tabular}{lccc}
    \toprule
    \textbf{Model} & \textbf{Text-centric} & \textbf{Vision-centric} & \textbf{Avg} \\
    \midrule
    Qwen2.5-VL 3B & \textit{23.97} & \textit{18.12} & \textit{21.05} \\
    Qwen2.5-VL 3B $\mathcal{D}_1$ & \textbf{\textit{62.08}} & \textit{44.80} & \textbf{\textit{52.49}} \\
    Qwen2.5-VL 3B $\mathcal{D}_2$ & \textit{51.18} & \textbf{\textit{50.50}} & \textit{50.84} \\
    \midrule
    Qwen2.5-VL 7B & \textit{37.75} & \textit{29.98} & \textit{33.87} \\
    Qwen2.5-VL 7B $\mathcal{D}_1$ & \textbf{\textit{74.22}} & \textit{53.77} & \textbf{\textit{64.00}} \\
    Qwen2.5-VL 7B $\mathcal{D}_2$ & \textit{63.42} & \textbf{\textit{60.40}} & \textit{61.91} \\
    \bottomrule
    \end{tabular}}
    \caption{Standard RL training on PGPS9K Results.}
    \label{tab:simple_rl_results_combined}
    \vspace{-0.05in}
\end{table}

\section{Mitigating Modality Gap}

In this section, we investigate an improved RL training recipe from two complementary perspectives to enhance the visual reasoning ability of MLLMs:

\noindent $\bullet$ \textbf{Data.} We explore two training strategies: \ding{172} \textit{mixed training}, which combines $\mathcal{D}_1$ (full-text) and $\mathcal{D}_2$ (partial-text) samples to expose models to both text- and vision-centric inputs; and \ding{173} \textit{curriculum training}, which first trains on $\mathcal{D}_1$ to consolidate reasoning under textual guidance, and then shifts to $\mathcal{D}_2$ to strengthen image-based reasoning and reduce shortcut reliance.  
    
\noindent $\bullet$ \textbf{Loss.} We introduce a KL-based self-distillation loss to align the model’s output distribution on $\mathcal{D}_2$ with that on $\mathcal{D}_1$, thereby preserving core reasoning ability while enhancing visual understanding.  

\subsection{Data Perspective}
\paragraph{Implementation.}We compare \textit{mixed training} and \textit{curriculum training}, matching the total training budget (details in Appendix~\ref{appendix:setup}). Curriculum training splits steps evenly: Stage 1 on $\mathcal{D}_1$, followed by Stage 2 on $\mathcal{D}_2$. 

\paragraph{Result.}As summarized in Table~\ref{tab:data_mixing_curriculum_results_combined}, curriculum training generally matches or surpasses mixed-data training in both in-distribution (PGPS9K) and out-of-distribution (MathVerse) evaluations.
Intuitively, Stage 1 on $\mathcal{D}_1$ consolidates general reasoning and solution formatting under rich textual guidance; Stage 2 on $\mathcal{D}_2$ then compels stronger visual grounding. This two-stage approach effectively improves both text-centric and vision-centric performance. 

\begin{table}[t]
    \centering
    \resizebox{\columnwidth}{!}{
    \begin{tabular}{lcccccc}
    \toprule
    & \multicolumn{3}{c}{\textbf{PGPS9K}} & \multicolumn{3}{c}{\textbf{MathVerse}} \\
    \cmidrule(r){2-4} \cmidrule(l){5-7}
    \textbf{Training Strategy} & \textbf{Text} & \textbf{Vision} & \textbf{Avg} & \textbf{Text} & \textbf{Vision} & \textbf{Avg} \\
    \midrule
    \multicolumn{7}{l}{\textit{Qwen2.5-VL 3B}} \\
    \quad Mixed training & 57.93 & 52.80 & 55.37 & 46.07 & 41.11 & 43.09 \\
    \quad Curriculum Stage 1 ($\mathcal{D}_1$) & 58.40 & 41.63 & 50.02 & \textbf{49.37} & 43.17 & 45.65 \\
    \quad Curriculum Stage 2 ($\mathcal{D}_1\!\rightarrow\!\mathcal{D}_2$) & \textbf{59.78} & \textbf{54.00} & \textbf{56.89} & 49.02 & \textbf{43.69} & \textbf{45.82} \\
    \midrule
    \multicolumn{7}{l}{\textit{Qwen2.5-VL 7B}} \\
    \quad Mixed training & 70.10 & 65.65 & 67.88 & 54.93 & 48.33 & 50.97 \\
    \quad Curriculum Stage 1 ($\mathcal{D}_1$) & \textbf{73.05} & 52.72 & 62.89 & 49.95 & 44.42 & 46.63 \\
    \quad Curriculum Stage 2 ($\mathcal{D}_1\!\rightarrow\!\mathcal{D}_2$) & 70.60 & \textbf{66.30} & \textbf{68.45} & \textbf{56.95} & \textbf{50.60} & \textbf{53.14} \\
    \bottomrule
    \end{tabular}}
    \caption{Data mixing and curriculum training results}
    \label{tab:data_mixing_curriculum_results_combined}
\end{table}

\subsection{Loss Perspective}

\begin{table}[t]
    \centering
    \resizebox{\columnwidth}{!}{
    \begin{tabular}{lcccccc}
    \toprule
    & \multicolumn{3}{c}{\textbf{PGPS9K}} & \multicolumn{3}{c}{\textbf{MathVerse}} \\
    \cmidrule(r){2-4} \cmidrule(l){5-7}
    \textbf{Training Strategy} & \textbf{Text} & \textbf{Vision} & \textbf{Avg} & \textbf{Text} & \textbf{Vision} & \textbf{Avg} \\
    \midrule
    \multicolumn{7}{l}{\textit{Qwen2.5-VL 3B}} \\
    \quad Plain RL on $\mathcal{D}_1$ & 58.40 & 41.63 & 50.02 & 49.37 & 43.17 & 45.65 \\
    \quad w/ KL & 58.10 & 45.47 & 51.79 & \textbf{49.57} & \textbf{43.67} & \textbf{46.03} \\
    \quad w/ KL + Curriculum  & \textbf{61.22} & \textbf{55.27} & \textbf{58.25} & 47.08 & 42.23 & 44.17 \\
    \midrule
    \multicolumn{7}{l}{\textit{Qwen2.5-VL 7B}} \\
    \quad Plain RL on $\mathcal{D}_1$ & 73.05 & 52.72 & 62.89 & 49.95 & 44.42 & 46.63 \\
    \quad w/ KL  & 73.28 & 54.30 & 63.79 & \textbf{57.00} & \textbf{48.79} & \textbf{52.07} \\
    \quad w/ KL + Curriculum & \textbf{73.42} & \textbf{67.87} & \textbf{70.65} & 53.76 & 48.55 & 50.63 \\
    \bottomrule
    \end{tabular}}
    \caption{Loss perspective results.}
    \label{tab:kl_loss_results_combined}
    \vspace{-0.1in}
\end{table}

\paragraph{Implementation.}We introduce a \emph{contrastive self-distillation KL loss} to transfer reasoning from full-text ($\mathcal D_1$) to partial-text ($\mathcal D_2$) inputs. Given paired prompts $(x^{(1)},x^{(2)})$, and a correct response $\hat{\mathbf y}$ sampled from $\pi_\theta(\cdot\mid x^{(1)})$, we align the partial-text distribution $p_t:=\pi_\theta(\cdot\mid \hat y_{<t},x^{(2)})$ with the frozen full-text distribution $q_t:=\operatorname{stopgrad}[\pi_\theta(\cdot\mid \hat y_{<t},x^{(1)})]$ via a time-averaged forward KL:
\begin{equation}
\label{eq:contrastive-kl}
\mathcal L_{\mathrm{cKL}}(\theta)
= \frac{1}{T}\sum_{t=1}^T \KL\big(p_t\,\|\,q_t\big).
\end{equation}

The forward KL encourages the model’s response distribution under partial-text inputs to \emph{cover} the high-confidence region of its own distribution under full-text inputs.
In practice, this KL loss is computed for all rollouts (without DAPO roll out batch group filtering) and added to the RL objective with weight $\alpha=0.01$, providing a dense learning signal and helping maintain the overall training loss optimization process stable. 
After the contrastive KL loss has stabilized, the model is further fine-tuned on $\mathcal{D}_2$ to enhance its visual reasoning ability.

\paragraph{Result.}We compare three training strategies in Table~\ref{tab:kl_loss_results_combined}: 
\ding{172} Plain RL on $\mathcal D_1$, 
\ding{173} with KL, i.e., adding the contrastive KL loss, and 
\ding{174} with\ KL + Curriculum, where the KL-trained model is subsequently fine-tuned on $\mathcal D_2$. 
From the in-distribution results on PGPS9K, both KL and KL + Curriculum consistently outperform the plain baseline, confirming that the KL term effectively transfers reasoning ability and stabilizes the optimization process. 
However, on the out-of-distribution dataset MathVerse, the improvements are less consistent, likely due to annotation and representation mismatches between the datasets. Specifically, PGPS9K provides explicit geometric cues, whereas MathVerse often omits such markings, weakening cross-domain transfer. We analyze this mismatch further in Appendix~\ref{appendix:annotation-difference}.
Overall, the KL loss enhances general reasoning ability, while the subsequent curriculum fine-tuning slightly degrades out-of-distribution performance, reflecting the impact of differing annotation styles across datasets. Please refer to Appendix~\ref{appendix:baseline_comparison} for more comparisons with baseline methods and Appendix~\ref{appendix:ablation_study} for additional ablation studies.

\section{Conclusion}

We present a systematic study on reducing the modality gap of MLLMs through reinforcement learning. Our experiments show that curriculum training effectively balances text-centric and vision-centric reasoning, and a KL-based self-distillation loss transfers reasoning competence from text-rich to vision-centric inputs. Together, these findings yield practical guidance: favor curriculum + contrastive KL to build MLLMs with stronger and more balanced visual reasoning capabilities.

\section*{Limitations}
Our proposed training recipe relies on constructing paired text-centric and vision-centric data, which currently leverages the detailed annotations of the PGPS9K geometry dataset. While we demonstrate that this method enhances performance across both domain-specific and general multimodal benchmarks (as shown in Appendix~\ref{appendix:baseline_comparison}), extending this data construction strategy to broader, less structured VQA tasks remains a challenge. Future work will explore automated methods to generate such training pairs for diverse domains, further validating the scalability of our approach.

\bibliography{custom}

\appendix
\section{Dataset Details}\label{appendix:dataset-details}

\paragraph{PGPS9K.}
PGPS9K is a large-scale, human-annotated dataset containing over $9000$ plain-geometry questions, split into $8000$ training and $1000$ test samples.

Each question comprises two components: a \emph{textual condition} and a \emph{question statement}.
The textual condition fully specifies the geometric construction—listing entities such as points, lines, and circles and relations including parallelism, perpendicularity, and congruence—while the question statement queries a particular geometric property (e.g., the length of a segment or the measure of an angle).

All figures include explicit annotations of entities and relations, which we refer to as \emph{full-condition images}.
All questions are free-form and admit a unique numerical answer.

\medskip
\noindent Based on these full-condition images, we define two dataset settings:
\begin{itemize}
    \item $\mathcal{D}_1$: \textbf{Full-condition question} $+$ \textbf{full-condition image}.  
    The textual condition fully specifies the geometry, resembling text-centric setups in typical VQA or reasoning datasets.
    \item $\mathcal{D}_2$: \textbf{Question only} $+$ \textbf{full-condition image}.  
    The textual condition is omitted, requiring the model to infer the geometry directly from the image, resulting in a more vision-centric and challenging setting.
\end{itemize}

\paragraph{MathVerse.}
We adopt the open-source subset \texttt{testmini} of the MathVerse dataset as our out-of-distribution evaluation benchmark. 

MathVerse contains two types of questions: multiple-choice and free-form questions, covering a broad range of visual-mathematical reasoning scenarios. 

The subset used in our experiments includes $778$ unique base questions, each instantiated into five variations: 
\emph{Text Dominant}, \emph{Text Lite}, \emph{Vision Intensive}, \emph{Vision Dominant}, and \emph{Vision Only}, yielding a total of $3890$ evaluation samples. These variations are designed to progressively reduce textual information while increasing dependence on visual cues, thus providing a systematic means of assessing the visual reasoning capability of multimodal large language models (MLLMs).  

For evaluation, we report three metrics: 
\ding{172} the average accuracy across all five variations, 
\ding{173} the average accuracy on the three vision-centric variations (\emph{Vision Intensive}, \emph{Vision Dominant}, and \emph{Vision Only}), and 
\ding{174} the average accuracy on the two text-centric variations (\emph{Text Dominant} and \emph{Text Lite}).

\section{Evaluation}
All results reported in this paper are obtained by sampling \textbf{four responses per question} (with a maximum response length of $4096$ tokens) and averaging Pass@1 across the samples.

For PGPS9K, we extract the final numerical answer from each response using regular expressions and compare it to the ground truth answer, which is also a number. A response is considered correct if the relative error is within $10^{-2}$.

For MathVerse, we also extract the final numerical answer from each response using regular expressions. However, since MathVerse includes both multiple-choice and free-form questions, 
we evaluate them differently: for multiple-choice questions, a response is correct if the extracted answer matches the correct choice; for free-form questions, a response is correct if the relative error is within $5\times 10^{-2}$.

\section{Complete Prompt}

\begin{tcolorbox}[left=1.2pt,right=1.2pt,top=1.2pt,bottom=1.2pt]
\small
\textbf{System Prompt}
\\
FIRST think about the reasoning process as an internal monologue and then provide the final answer. The reasoning process MUST BE enclosed within <think> </think> tags. The final answer MUST BE put in \verb|\boxed{<final answer>}|.

\textbf{Input Image Example:}
\\
\includegraphics[width=0.3\textwidth]{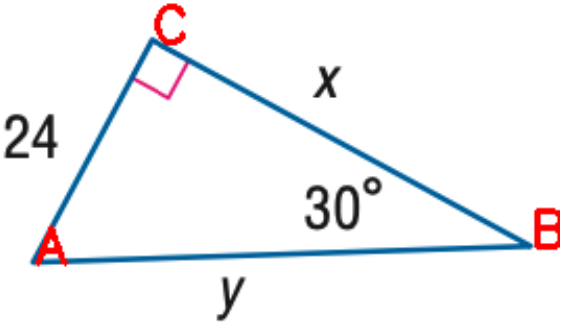}
\\
\textbf{Prompt For Text-centric Task}

In this problem, \(CB \perp CA\) at \(C\), \(AC = 24\), \(BC = x\), \(AB = y\), and \(m\angle CBA = 30^\circ\). Based on these conditions, answer the question: Find \(y\).
\\
\textbf{Prompt For Vision-centric Task}
\\
Based on the conditions in the image, answer the question: Find \(y\).
\\
\end{tcolorbox}

\section{Dynamic sAmpling Policy Optimization(DAPO) Overview}
DAPO\cite{yu2025dapo} serves as the base RL training method in our pipeline. For clarity, relative to the standard GRPO framework, DAPO introduces the following modifications:
\begin{itemize}
    \item \textbf{Clip-Higher.} DAPO decouples the lower and upper clipping ranges $(\epsilon_{\text{low}}, \epsilon_{\text{high}})$ of the importance-sampling ratios and enlarges the upper bound. This mitigates entropy collapse and improves exploration by allowing greater flexibility for low-probability exploration tokens during policy updates.
    \item \textbf{Dynamic sampling.} DAPO filters out prompts whose response groups are uniformly correct or uniformly incorrect, as these prompts do not generate meaningful gradients under GRPO. Since this filtering reduces the effective batch size, DAPO oversamples prompts that yield non-trivial advantages to maintain a sufficient number of gradient-contributing samples. This increases the density of useful training signals and improves the stability of GRPO-style updates.
    \item \textbf{Token-level policy gradient loss.} In vanilla GRPO, all token log-probability terms within a response are averaged first, and this averaged value becomes the sample’s sole contribution to the update. Consequently, samples of different lengths receive equal weight, which dilutes the gradients of long chain-of-thought responses and overemphasizes short responses. DAPO eliminates this sample-level averaging and aggregates losses at the token level. Each token therefore contributes directly to the optimization objective, preventing long responses from being under-weighted and enabling finer-grained credit assignment. This produces more stable updates in long sequence RL and ensures that informative reasoning steps and error-inducing tokens are appropriately reflected in the gradient signal.
    \item \textbf{Overlong reward shaping.} To address instability caused by excessively long or truncated responses, DAPO applies a length-aware penalty known as Soft Overlong Punishment. When a response exceeds a predefined maximum length, DAPO introduces a punishment interval in which the penalty increases smoothly with response length. Shorter responses receive no penalty, while severely overlong responses receive a fixed maximum penalty. This penalty is added to the rule-based correctness reward to discourage unnecessarily long outputs while preserving valid reasoning content. This mechanism reduces reward noise from truncation and prevents models from exploiting longer outputs to manipulate rewards.
\end{itemize}

\section{Training Setup}\label{appendix:setup}

All training for RL and ablations is conducted on the PGPS9K training set, and evaluation is performed on the PGPS9K test split and the MathVerse testmini subset. All reinforcement learning experiments are conducted with DAPO under the following configuration: 

\textbf{Clipping ratios.} Lower and upper clipping thresholds are set to $0.2$ and $0.28$, with an additional coefficient $c=10.0$ for actor-critic stability.  

\textbf{Overlong responses.} To handle long generations, we use a buffer length of $1024$, enable buffer control, and apply a penalty factor of $1.0$ when responses exceed this limit.

\textbf{Training configuration.} Batch size is $512$ and mini-batch size is $128$, with maximum prompt length of $1024$ and maximum response length of $4096$. The learning rate is fixed at $1 \times 10^{-6}$.  

\textbf{Stopping criterion.} Unless otherwise noted, training is stopped once the DAPO parameter \texttt{num\_gen\_batches} reaches $10$, which means that $10$ rollout steps are required to accumulate one gradient update.  

\textbf{Models.} We use Qwen2.5-VL 3B and 7B as our base models, which are open-source MLLMs with strong performance on visual reasoning tasks.

\textbf{Computing Infrastructure.} All experiments are conducted on 8 H100 GPUs with 80GB memory each. Each training run takes approximately 24 hours for Qwen2.5-VL 3B and 48 hours for Qwen2.5-VL 7B.

These settings are used consistently across all experiments to ensure comparability.

\section{Ablation on RL under a Fixed Training Distribution}\label{appendix:ablation_study}

\begin{table}[t]
\centering
\small
\resizebox{\columnwidth}{!}{
\begin{tabular}{lcccc}
\toprule
Model & \textbf{Text-centric} ($\mathcal{D}_1$) & \textbf{Vision-centric} ($\mathcal{D}_2$) & \textbf{Text-only} & \textbf{Gap} \\
\midrule
Qwen3B Base Model & 23.97 & 18.12 & 16.10 & $-$2.02 \\
Qwen3B Plain RL 240    & 56.50 & \textbf{46.45} & 46.82 & 0.37 \\
Qwen3B Plain RL 320    & \textbf{62.08} & 44.80 & \textbf{48.70} & 3.90 \\
\midrule
Qwen7B Base Model & 37.75 & 29.98 & 32.95 & 2.97 \\
Qwen7B Plain RL 200    & 72.68 & 53.17 & 60.02 & 6.85 \\
Qwen7B Plain RL 280    & \textbf{74.22} & \textbf{53.77} & \textbf{62.10} & 8.33 \\
\bottomrule
\end{tabular}
}
\caption{Ablation results under a fixed training distribution. All models are trained exclusively on the Text-centric ($\mathcal{D}_1$) subset. The gap (Text-only $-$ Vision-centric) increases consistently with additional RL steps, indicating that reinforcement learning disproportionately strengthens text-based reasoning.}
\label{tab:rl_modality_ablation}
\end{table}

To determine whether the observed modality gap is caused by RL training itself, rather than by a train--test distribution mismatch induced by splits of PGPS9K, we conduct a controlled evaluation in which the training distribution is strictly fixed. All models are trained exclusively on the \textbf{Text-centric ($\mathcal{D}_1$)} subset of PGPS9K, where both full textual descriptions and images are available. No modality-ablated subsets are used during training, and all RL updates are derived solely from this data.

After training, the same model checkpoints are evaluated under three inference settings: (1) \textbf{Text-centric ($\mathcal{D}_1$)}, which includes both complete text and image inputs; (2) \textbf{Vision-centric ($\mathcal{D}_2$)}, where textual descriptions are omitted, leaving only images and questions; and (3) \textbf{Text-only}, where images are withheld, providing only textual descriptions and questions. Since the training data distribution is identical across all models, performance differences across evaluation settings reflect changes in inference behavior rather than distribution mismatch.

Table~\ref{tab:rl_modality_ablation} reports the results for Qwen models of different sizes and RL training steps. As RL progresses, performance improves consistently across all evaluation settings; however, the improvement is not uniform. Specifically, performance in the \textbf{Text-only} setting increases more rapidly than in the \textbf{Vision-centric} setting. For both 3B and 7B models, later-stage RL checkpoints exhibit higher accuracy in the Text-only setting compared to the Vision-centric setting, even though the models were never explicitly trained on text-only data.

This asymmetric improvement leads to a widening gap between text-only and vision-centric performance as RL steps increase. In contrast, base models exhibit a much smaller gap, sometimes performing comparably or even better in the Vision-centric setting. Because all models are trained on the same Text-centric ($\mathcal{D}_1$) dataset, these results indicate that reinforcement learning intrinsically biases optimization toward text-dominant reasoning, thereby amplifying the modality gap even when the training distribution is held constant.

\section{Comparison With Other Baseline Methods and General Benchmarks}\label{appendix:baseline_comparison}

To provide a direct comparison with existing baseline methods and assess the effectiveness of our proposed training strategy, we include two widely used baselines for multimodal reasoning. The first is an in-context learning baseline following MMICL~\cite{zhao2024mmiclempoweringvisionlanguagemodel}, where four multimodal examples are provided at inference time. The second is Perception-Aware Policy Optimization (PAPO)~\cite{wang2025perception}, a reinforcement learning method designed to enhance multimodal perception during training. All models are trained exclusively on PGPS9K train set. All benchmarks discussed below are used solely for evaluation and comparison across different training strategies. All baselines are evaluated under the same inference protocol as our method.

We first compare these baselines on PGPS9K under both text and vision evaluation settings. Results are shown in Table~\ref{tab:pgps9k_baseline}. Across both 3B and 7B model scales, our KL+Curriculum strategy outperforms generalized base models, MMICL-style in-context learning, and PAPO. 

\begin{table}[t]
\centering
\resizebox{\columnwidth}{!}{
\begin{tabular}{lccc}
\toprule
Model & \textbf{Text-centric} & \textbf{Vision-centric} & \textbf{Avg} \\
\midrule
Qwen 2.5 7B Base Model & 37.75 & 29.98 & 33.87 \\
Qwen 2.5 7B MMICL & 37.18 & 30.70 & 33.94 \\
Qwen 2.5 7B PAPO & 45.87 & 37.70 & 41.79 \\
Qwen 2.5 7B KL+Curriculum (Ours) & \textbf{73.42} & \textbf{67.87} & \textbf{70.65} \\
\midrule
Qwen 2.5 3B Base Model & 23.97 & 18.12 & 21.05 \\
Qwen 2.5 3B MMICL & 24.07 & 20.10 & 22.14 \\
Qwen 2.5 3B PAPO & 39.87 & 33.67 & 36.77 \\
Qwen 2.5 3B KL+Curriculum (Ours) & \textbf{61.22} & \textbf{55.27} & \textbf{58.25} \\
\bottomrule
\end{tabular}}
\caption{Comparison with baseline methods on PGPS9K under Text-centric and Vision-centric evaluation settings.}
\label{tab:pgps9k_baseline}
\end{table}

We further evaluate the same set of models on MathVerse, which tests multimodal mathematical reasoning beyond the training distribution. Results in Table~\ref{tab:mathverse_baseline} show that our method consistently achieves the best average performance across both text and vision settings at different model scales, demonstrating that the gains are not limited to PGPS9K.

\begin{table}[t]
\centering
\resizebox{\columnwidth}{!}{
\begin{tabular}{lccc}
\toprule
Model & \textbf{Text-centric} & \textbf{Vision} & \textbf{Avg} \\
\midrule
Qwen 2.5 7B Base Model & 55.01 & 45.76 & 51.10 \\
Qwen 2.5 7B MMICL & 55.92 & 48.15 & 51.26 \\
Qwen 2.5 7B PAPO & 52.73 & 45.72 & 48.52 \\
Qwen 2.5 7B KL (Ours) & \textbf{57.00} & \textbf{48.79} & \textbf{52.07} \\
\midrule
Qwen 2.5 3B Base Model & 35.68 & 28.66 & 31.47 \\
Qwen 2.5 3B MMICL & 43.80 & 36.21 & 39.24 \\
Qwen 2.5 3B PAPO & 47.87 & 40.57 & 43.49 \\
Qwen 2.5 3B KL (Ours) & \textbf{48.19} & \textbf{41.58} & \textbf{44.23} \\
\bottomrule
\end{tabular}}
\caption{Results on MathVerse under Text-centric and Vision-centric evaluation settings.}
\label{tab:mathverse_baseline}
\end{table}

Finally, we evaluate the same set of models on two widely used general benchmarks, MATH500~\cite{lightman2023lets} and MMSTAR~\cite{chen2024rightwayevaluatinglarge}. As shown in Table~\ref{tab:general_benchmark}, training on PGPS9K with different optimization strategies does not degrade performance on general QA and VQA benchmarks. In particular, our KL-regularized curriculum training achieves performance comparable to existing baselines across both benchmarks, indicating that geometry-focused training on PGPS9K does not adversely affect the model’s general reasoning and perception capabilities.

\begin{table}[t]
\centering
\resizebox{\columnwidth}{!}{
\begin{tabular}{lcc}
\toprule
Model & \textbf{MATH500} & \textbf{MMSTAR} \\
\midrule
Qwen 2.5 7B Base Model & 65.80 & 60.47 \\
Qwen 2.5 7B PAPO & 59.20 & \textbf{63.00} \\
Qwen 2.5 7B MMICL & \textbf{66.40} & 61.00 \\
Qwen 2.5 7B KL+Curriculum (Ours) & 62.20 & 61.87 \\
\midrule
Qwen 2.5 3B Base Model & 54.60 & 50.13 \\
Qwen 2.5 3B PAPO & 61.00 & 54.40 \\
Qwen 2.5 3B MMICL & \textbf{64.60} & 49.93 \\
Qwen 2.5 3B KL+Curriculum (Ours) & 59.40 & \textbf{54.80} \\
\bottomrule
\end{tabular}}
\caption{Results on general benchmarks MATH500 and MMSTAR.}
\label{tab:general_benchmark}
\end{table}

\section{Annotation Difference in Two Dataset}\label{appendix:annotation-difference}
\begin{figure}[t]
    \centering
    \includegraphics[width=1.0\columnwidth]{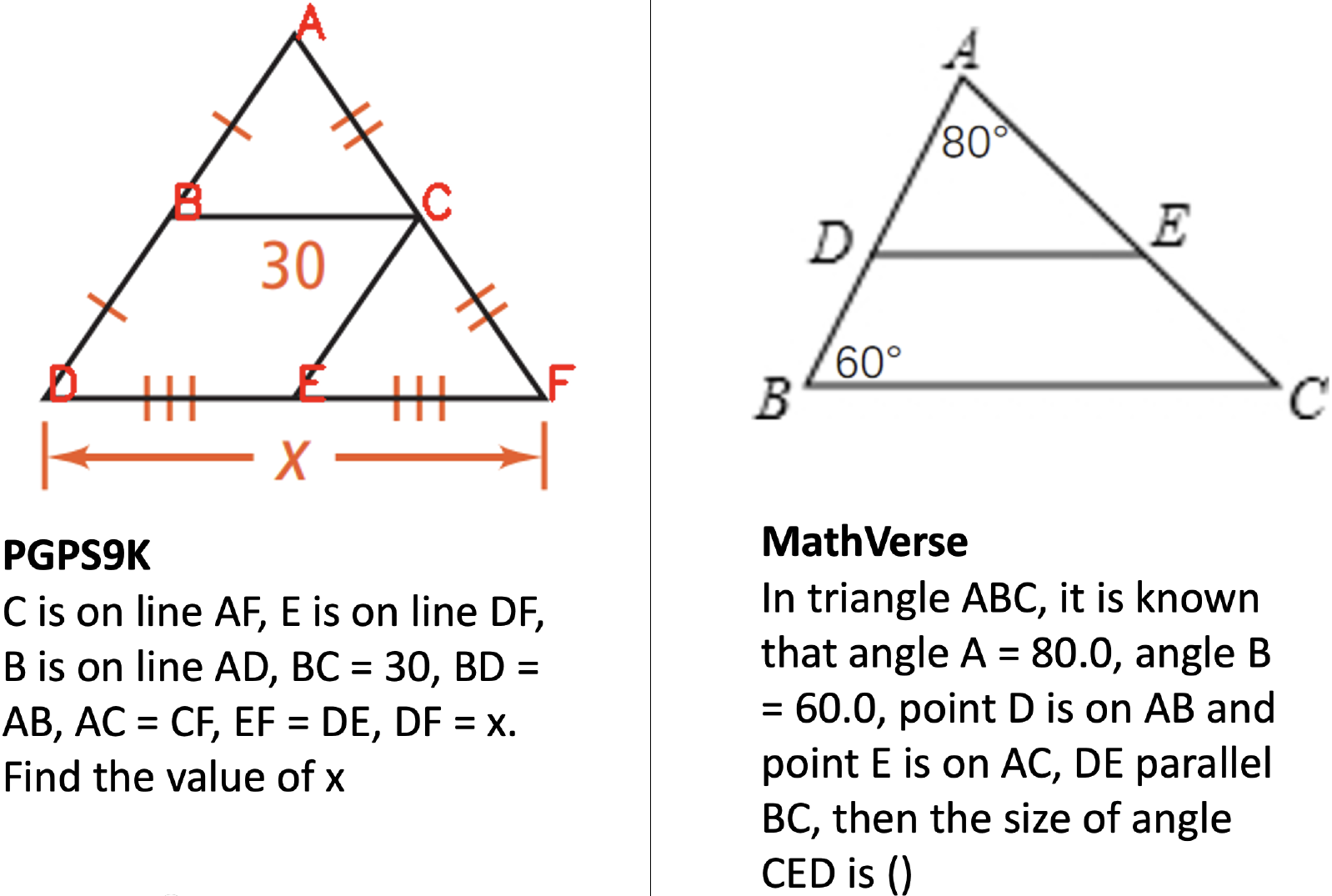}
    \caption{Annotation style mismatch between PGPS9K and MathVerse. PGPS9K diagrams explicitly mark key geometric relations—parallelism and equality of segments/angles—whereas MathVerse omits such markings. Models trained on PGPS9K may over-rely on these visual tags and struggle to infer relations on MathVerse, weakening out-of-distribution generalization.}
    \label{fig:annotation_difference}
\end{figure}
One key reason models trained on PGPS9K sometimes underperform on MathVerse is a mismatch in \emph{annotation style}. As shown in Figure~\ref{fig:annotation_difference}, PGPS9K explicitly marks geometric relations on the diagram—most notably \ding{172} segment parallelism and \ding{173} equivalence relations between segments and angles(e.g., equal-length segments and equal/corresponding/alternate angles). By contrast, MathVerse does \emph{not} provide these markings. In several MathVerse settings, the model must infer these relations directly from the geometry without explicit visual tags, so a model trained on PGPS9K’s fully annotated figures can overfit to those cues and exhibit weaker out-of-distribution generalization on MathVerse.

\section{Artifacts License}
Our training codes primarily build upon the open-source training framework verl~\cite{sheng2025hybridflow}, which is licensed under the Apache-2.0 License.

All source code developed for this work will be released under the Apache-2.0 License, which permits both research and commercial use, along with modifications and distribution.

The two datasets used in this work, MathVerse~\cite{zhang2024mathverse} and PGPS9K~\cite{zhang2023multi} 
 are licensed under MIT License, which allows for free use, modification, and distribution.

The Qwen2.5-VL series models~\cite{Qwen2.5-VL} are released under the Apache-2.0 License, which permits both research and commercial use, along with modifications and distribution.

\end{document}